\documentclass[10pt,conference]{IEEEtran}

\usepackage[utf8]{inputenc} % Allow UTF-8 input
\usepackage[T1]{fontenc}    % Use 8-bit T1 fonts
\usepackage{url}            % Simple URL typesetting
\usepackage{booktabs}       % Professional-quality tables
\usepackage{amsfonts}       % Blackboard math symbols
\usepackage{nicefrac}       % Compact symbols for 1/2, etc.
\usepackage{microtype}      % Microtypography
\usepackage{xcolor}         % Colors
\usepackage{soul}           % Text highlighting
\usepackage{graphicx}       % Figures
\usepackage{float}          % Float management
\usepackage{multicol}       % Multiple columns
\usepackage{multirow}       % Multirow in tables
\usepackage{cite}           % IEEE citation formatting
\usepackage{array}          % Table improvements
\usepackage{makecell}       % Enhanced table cells
\usepackage{amsmath}        % Math support
\usepackage{tabularx}       % Advanced table support
\usepackage{colortbl}       % Colored tables
\usepackage{pgfplots}       % PGFPlots for graphs
\usepackage{pgfplotstable}  % PGF Table support
\usepackage{amssymb}
\usepackage{mathtools}
\pgfplotsset{compat=1.17}

\usepackage{tikz}           % TikZ for graphics
\usetikzlibrary{positioning}
\usetikzlibrary{shapes.geometric, decorations.pathreplacing} 
\tikzset{>=latex}

\usepackage{siunitx}
\usepackage[outline]{contour} % glow around text
\contourlength{1.2pt}
\usetikzlibrary{3d} % for canvas
\usetikzlibrary{shadows.blur}
\usetikzlibrary{fadings}
\usetikzlibrary{decorations.text} % for text along path
\usetikzlibrary{shapes.misc}
\usetikzlibrary{pgfplots.statistics}
\usetikzlibrary{fit}

\tikzset{>=latex}
\tikzstyle{xlab}=[below=-1,scale=0.85]
\tikzstyle{ylab}=[left=-1,scale=0.85]

\colorlet{mydarkblue}{blue!40!black}
\colorlet{mylightblue}{mydarkblue!12} %blue!70!black!20
\colorlet{myred}{red!80!black}
\colorlet{mydarkred}{red!50!black}
\colorlet{mylightred}{mydarkred!12}
\colorlet{mydarkgreen}{green!30!black}
\colorlet{mylightgreen}{mydarkgreen!12}
\colorlet{myorange}{orange!63!black}
\colorlet{mylightorange}{orange!80!black!12}

\usetikzlibrary{patterns} % for hatched patterns
\def\hatchsize{4pt}
\makeatletter
\pgfdeclarepatternformonly{myhatch}{%
  \pgfqpoint{-1pt}{-1pt}}{\pgfqpoint{\hatchsize}{\hatchsize}}{\pgfqpoint{\hatchsize}{\hatchsize}
}{%
  \pgfsetlinewidth{0.3pt}
  \pgfpathmoveto{\pgfqpoint{0pt}{\hatchsize}}
  \pgfpathlineto{\pgfqpoint{\hatchsize}{0pt}}
  \pgfusepath{stroke}
}
\makeatother

\tikzfading[
  name=fade out,
  inner color=transparent!0,
  outer color=transparent!100
]
\tikzfading[
  name=fade right,
  left color=transparent!0,
  right color=transparent!100
]

\usetikzlibrary{shadings}
\makeatletter
\pgfdeclareradialshading[tikz@radial@inner,tikz@radial@outer]%
  {myradial}{\pgfpointorigin}{% \pgfqpoint{-50bp}{-50bp}}{%
    color(0bp)=(pgftransparent!0);%
    color(20bp)=(pgftransparent!90);%
    color(30bp)=(pgftransparent!93);%
    color(40bp)=(pgftransparent!97);%
    color(50bp)=(pgftransparent!100)%
  }
\makeatother
\pgfdeclarefading{myradial}{\pgfuseshading{myradial}}%

\def\minwidth{5cm}

\tikzstyle{pooling} = [draw, rectangle, minimum width=\minwidth, minimum height=.75cm, fill=red!20, text centered, anchor=north]

\tikzstyle{activation} = [draw, rectangle, minimum width=\minwidth, minimum height=.75cm, fill=gray!20, text centered, anchor=north]

\tikzstyle{container} = [draw, rectangle, minimum width=\minwidth, minimum height=.75cm, fill=yellow!20, text centered, anchor=north]

\tikzstyle{norm} = [draw, rectangle, minimum width=\minwidth, minimum height=.75cm, fill=blue!20, text centered, anchor=north]

\tikzstyle{conv} = [draw, rectangle, minimum width=\minwidth, minimum height=.75cm, fill=purple!20, text centered, anchor=north]

\tikzstyle{mul} = [draw, rectangle, minimum width=\minwidth, minimum height=.75cm, fill=green!20, text centered, anchor=north]

\tikzstyle{downsample} = [draw, trapezium, trapezium left angle=120, trapezium right angle=120, minimum width = \minwidth, minimum height = .5cm, fill = orange!20, text centered, anchor =north]

\tikzstyle{upsample} = [draw, trapezium, trapezium left angle= 60, trapezium right angle=60, minimum width = \minwidth, minimum height = .5cm, fill = orange!20, text centered, anchor =north]

\tikzstyle{linear} = [draw, rectangle, minimum width=\minwidth, minimum height=.75cm, fill=orange!20, text centered, anchor=north]

\makeatletter
\def\maketag@@@#1{\hbox{\m@th\normalfont\normalsize#1}}
\makeatother

\title{Contrastive Self-Supervised Network Intrusion Detection using Augmented Negative Pairs}

\author{
    \IEEEauthorblockN{
        Jack Wilkie\IEEEauthorrefmark{1}, 
        Hanan Hindy\IEEEauthorrefmark{2}, 
        Christos Tachtatzis\IEEEauthorrefmark{1}, 
        Robert Atkinson\IEEEauthorrefmark{1}
    }
    \IEEEauthorblockA{\IEEEauthorrefmark{1}University of Strathclyde, Glasgow, UK}
    \IEEEauthorblockA{\IEEEauthorrefmark{2}Ain Shams University, Cairo, Egypt}
}

\begin{document}

\IEEEoverridecommandlockouts

\IEEEpubid{\begin{minipage}{\textwidth}\ \vspace{1cm}\\[10pt]
\centering
\copyright~2025 IEEE. Personal use of this material is permitted. 
Permission from IEEE must be obtained for all other uses, in any current or future media, 
including reprinting/republishing this material for advertising or promotional purposes, 
creating new collective works, for resale or redistribution to servers or lists, 
or reuse of any copyrighted component of this work in other works.
DOI: \url{https://doi.org/10.1109/CSR64739.2025.11129979}
Link to IEEE Xplore: \url{https://ieeexplore.ieee.org/document/11129979}
\end{minipage}}

\maketitle

\begin{abstract}
  Network intrusion detection remains a critical challenge in cybersecurity. While supervised machine learning models achieve state-of-the-art performance, their reliance on large labelled datasets makes them impractical for many real-world applications. Anomaly detection methods, which train exclusively on benign traffic to identify malicious activity, suffer from high false positive rates, limiting their usability. Recently, self-supervised learning techniques have demonstrated improved performance with lower false positive rates by learning discriminative latent representations of benign traffic. In particular, contrastive self-supervised models achieve this by minimising the distance between similar (positive) views of benign traffic while maximising it between dissimilar (negative) views. Existing approaches generate positive views through data augmentation and treat other samples as negative. In contrast, this work introduces Contrastive Learning using Augmented Negative pairs~(CLAN), a novel paradigm for network intrusion detection where augmented samples are treated as negative views—representing potentially malicious distributions—while other benign samples serve as positive views. This approach enhances both classification accuracy and inference efficiency after pretraining on benign traffic. Experimental evaluation on the Lycos2017 dataset demonstrates that the proposed method surpasses existing self-supervised and anomaly detection techniques in a binary classification task. Furthermore, when fine-tuned on a limited labelled dataset, the proposed approach achieves superior multi-class classification performance compared to existing self-supervised models.

  \end{abstract}


\section{Introduction}
\label{sec:introduction} 
\IEEEPARstart{M}{achine} Learning~(ML) models have become the leading approach in Network Intrusion Detection Systems~(NIDS). These models have achieved State-of-the-art~(SOTA) performance~\cite{Hindy_2020} and, unlike traditional methods, do not require experts to painstakingly identify patterns, known as signatures, or develop classification rules for known intrusions. However, ML models traditionally require large labelled datasets, containing many examples of each class the system aims to classify, to be effective. Unfortunately, in network intrusion detection acquiring such datasets is often non-trivial, necessitating that experts identify and label numerous instances of malicious network flows. To exacerbate this issue, traffic taken from other networks, either real or simulated, does not generalise well to new networks~\cite{Layeghy_2023}. This leaves ML-based NIDS non-implementable in newly established networks where there are no labelled instances of malicious traffic. Furthermore, fully trained ML classifiers struggle to detect novel ``zero-day'' intrusions for which there are no labelled instances.

One solution to this problem is to exploit the abundance of benign network flows through the application of anomaly  detection algorithms, such as Support Vector Machines~(SVMs)~\cite{Hanan_svm1} and autoencoders~\cite{10.1007/978-3-319-13563-2_27}. These algorithms learn the distribution of benign network traffic during training and flag non-conforming traffic as malicious during inference. While widely researched, the lack of malicious samples during training results in traditional anomaly detectors exhibiting false positive rates that are too high for practical deployment.

Recently, Self-Supervised Learning~(SSL) has emerged as a promising approach for NIDS, enabling models to learn meaningful latent representations from unlabelled data~\cite{balestriero2023cookbookselfsupervisedlearning,conflow,9935282,10619725,byol2}. This label independence allows it to be applied to NIDS to learn meaningful representations of network traffic from only benign flows~\cite{non_contrastive_ssl}. Contrastive learning is one such method, where models are trained by minimising a distance metric between similar (positive) pairs of samples while simultaneously maximising it between dissimilar (negative) pairs. This is typically achieved by generating augmented views of the same sample, which are treated as positive pairs, resulting in a latent representation which is robust to the chosen perturbations. Several works have successfully leveraged contrastive SSL to improve intrusion detection performance~\cite{conflow,9935282,10619725}, however, existing SSL approaches learn a distinct distribution for each sample and its augmented views in latent space. This limits their ability to model benign traffic wholistically and introduces challenges in distinguishing between benign and malicious traffic effectively.

\IEEEpubidadjcol

This work proposes Contrastive Learning using Augmented Negative Pairs~(CLAN), which presents a change in paradigm: instead of treating augmented samples as positive pairs, they are treated as negative. It is shown that this change results in the model learning a fundamentally different latent representation of the data: while existing approaches learn a distinct latent distribution for each sample; CLAN instead learns a single distribution of benign traffic. Not only does this allow for more efficient inference, but by learning the distribution of benign traffic wholistically CLAN achieves improved performance when both deployed as an anomaly detector or fine-tuned on a limited dataset to perform multi-class classification. The core contributions of this work can be summarised as follows:

\begin{enumerate}
    \item A contrastive SSL framework is proposed for NIDS. In contrast to traditional approaches, where samples and their augmented versions are treated as positive pairs, this work instead aims to model the class level distribution of benign traffic by viewing augmented samples as belonging to another, potentially malicious, distribution. Thus, they are treated as negative pairs.
  
    \item The framework is extended to perform binary classification without fine-tuning allowing it to function as an anomaly detector. Experimental results show the proposed approach outperforms existing anomaly detection and SSL algorithms.
    
    \item  It is shown that the priors learned by pretraining can be exploited for supervised classification, allowing the proposed model to be performative when fine-tuned on a limited quantity of labelled samples. It is experimentally shown to outperform existing SSL approaches in this setting.
  \end{enumerate}
  
This work is arranged as follows: Section~\ref{sec:related_works} begins by describing work related to the proposed approach, including NIDS, anomaly detection, and SSL. Section~\ref{sec:proposed_approach} introduces the proposed framework and extends it to anomaly detection, with Section~\ref{sec:contrastive_comparison} providing an in-depth comparison to existing SSL approaches. Experimental evaluation is preformed in Section~\ref{sec:results} where the model is compared to existing models in anomaly detection and fine-tuned performance. Finally, Section~\ref{sec:conclusions} concludes this work with relevant discussion and conclusions based on the experimental findings in its preceding section. 

\section{Related Works}
\label{sec:related_works}
This section details work the background literature related to the CLAN. Section~\ref{subsec:related_nids} introduces NIDS and the challenges in building an adequate training dataset, with Section~\ref{subsec:related_anomaly} summarising attempts at anomaly detection based solutions. Finally, Section~\ref{subsec:related_ssl} details self-supervised learning and its application to NIDS.

\subsection{Network Intrusion Detection Systems}
\label{subsec:related_nids}

Network Intrusion Detection Systems are used to monitor network traffic to prevent unauthorised access or attacks. Traditional NIDS relied on signature-based detection, where experts manually crafted features and classification rules to uniquely identify each attack class. While these systems achieved high precision, they faced significant scalability challenges, as extending them to accommodate the exponentially growing number of attacks required substantial human effort. Additionally, signature-based systems are ineffective against novel intrusions, leaving networks vulnerable to zero-day attacks.

To address these limitations, ML has emerged as the dominant approach for reducing the manual effort required to develop effective NIDS. These models operate by learning statistical patterns from historical network traffic data to classify and detect attacks. Gradient-free approaches such as decision trees and Support Vector Machines~(SVMs)~\cite{Hanan_svm1} were initially employed to partition input features into regions of benign and malicious traffic. Since then, deep learning architectures such as multi-layer perceptrons~(MLPs), Convolutional Neural Networks~(CNNs)~\cite{Xiao2019} and Long Short-Term Memory~(LSTM) networks~\cite{CNN_LTSM} have been achieved SOTA performance by training parameterised models to learn non-linear decision boundaries.

Despite their success, ML-based classifiers require large amounts of labelled training data, which can be difficult and expensive to obtain. Furthermore, their performance deteriorates significantly when confronted with a zero-day attack. 

\subsection{Anomaly Detection}
\label{subsec:related_anomaly}

Anomaly detection-based approaches have been employed to mitigate the challenges associated with acquiring labelled datasets for training ML-based NIDS. These methods learn the distribution of exclusively benign network traffic during training and flag non-conforming traffic as malicious. Since they do not rely on labelled malicious samples, anomaly detection techniques are able to detect all attacks, both known and zero-day, equally well.

Statistical anomaly detection methods rely on distance metrics or probabilistic models to distinguish between normal and anomalous traffic. Distance-based approaches compute the centroid of benign traffic in the training dataset and classify test samples based on their Minkowski distance from this centroid~\cite{doi:10.1137/1.9781611976236.18}. This approach has been extended to alternative distance metrics, such as the Frobenius and Grassmannian distance measures~\cite{10.1007/978-3-319-70087-8_59}. Other statistical methods, including local outlier factor and nearest-neighbour distance-based techniques, detect anomalies by examining the density of local neighbourhoods rather than relying on global statistics~\cite{electronics9061022}. Another category of statistical methods extends traditional discriminative models to anomaly detection, such as one-class SVMs~\cite{Hanan_svm1} and isolation forests~(IF)~\cite{10040395}, which adapt SVMs and tree-based models, respectively.

More recently, deep learning-based anomaly detection has gained popularity due to its ability to learn more complex, non-linear decision boundaries than statistical methods. Autoencoders~(AE)~\cite{10.1007/978-3-319-13563-2_27} are a widely used approach where the model is trained to reconstruct input samples from a compressed latent representation. The reconstruction error is then used as an anomaly score, with higher errors indicating deviations from normal traffic. Variants such as sparse autoencoders~\cite{9239385}, deep unsupervised anomaly detection~(DUAD)~\cite{li2021deep}, and DAE-LR~\cite{nkashama2024deeplearningnetworkanomaly} build on this principle with various modifications. Hybrid approaches, such as AutoSVM~\cite{8463474} and Deep Support Vector Data Descriptor~(Deep SVDD)~\cite{pmlr-v80-ruff18a}, and Deep Gaussian Mixture Models~(DAGMM)~\cite{purohit2020deepautoencodinggmmbasedunsupervised} integrate deep learning methods with statistical techniques to enhance anomaly detection performance. Finally, generative approaches such as GANs~\cite{9219561} and variational autoencoders~\cite{9017945} have been used to train a classifier to identify artificially generated samples as malicious distributions.

While anomaly detectors enable classifiers trained on only benign traffic, the lack of malicious examples at train time results in them having false positive rate too high to be used in practice~\cite{10.1007/978-3-319-13563-2_27}.

\subsection{Self-Supervised Learning}
\label{subsec:related_ssl}

Self-supervised learning has emerged as a promising approach for future NIDS. SSL enables models to learn semantic representations from unlabelled data by leveraging pretext tasks such as masked autoencoding~\cite{touvron2023llamaopenefficientfoundation} and restoration~\cite{DBLP:journals/corr/ZhangIE16}. Exploiting this could allow NIDS to be trained on only benign data whilst maintaining an acceptable false positive rate.

One branch of SSL learns semantic representations by minimising a distance metric between positive views of a sample. These views are often generated through a series of augmentations, with the resultant model learning a latent representation of the data which is invariant to the chosen augmentations. However, this can often lead to the degenerative solution where all inputs are mapped to a point, known as dimensional collapse. Contrastive learning methods such as SimCLR~\cite{chen2020simpleframeworkcontrastivelearning} and infoNCE~\cite{rusak2024infonceidentifyinggaptheory} attempt to prevent dimensional collapse by simultaneously maximising the distance between an input sample and its negative views, most often other samples. Similarly, knowledge distillation methods such as BYOL~\cite{grill2020bootstraplatentnewapproach} and SimSiam~\cite{chen2020exploringsimplesiameserepresentation} prevent dimensional collapse using architectural tricks such as momentum encoders. Finally, the canonical correlation family of techniques, including models such as VICReg~\cite{bardes2022vicregvarianceinvariancecovarianceregularizationselfsupervised} and Barlow Twins~\cite{zbontar2021barlowtwinsselfsupervisedlearning} prevent dimensional collapse by maximising a lower bound on the information in the output matrices.

Several existing works have applied SSL to NIDS, with contrastive learning being the most common approach. For instance, SSCL-IDS~\cite{10619725} minimises the distance between positive pairs generated using cutmix while maximising the distance between other samples. Similarly, Conflow~\cite{conflow} and CLDNN~\cite{9935282} generate positive pairs using different dropout masks and feature masking, respectively. Research has also explored the use of knowledge distillation methods like BYOL and SimSiam, as well as canonical correlation-based models such as VICReg and Barlow Twins, as these approaches have been shown in other domains to train effectively with lower batch sizes~\cite{non_contrastive_ssl}. These methods have demonstrated promising results by outperforming traditional anomaly detection techniques.

While SSL-based NIDS has shown promise, existing methods use augmentations to generate positive pairs. This work instead treats them as negative pairs, whilst using other samples as positive pairs. It is shown that this allows the model to explicitly model the distribution of benign traffic resulting in improved performance and more efficient inference.

\section{Proposed Approach}
\label{sec:proposed_approach}

\begin{figure*}[t]
    \centering
    \resizebox{\linewidth}{!}{
        \begin{tikzpicture}[
    node distance=1.5cm and 2cm,
    box/.style={draw, rectangle, minimum width=2cm, minimum height=1cm, align=center, inner sep=5pt},
    arrow/.style={->, thick},
    image/.style={inner sep=0.pt, anchor=center},
    ] 
    \newcommand{\labelspacing}{0.2cm}

    %  -- Input
    \node[box, fill=red!20] (x00) {$x_0$};
    \node[box, fill=orange!20, below=0cm of x00] (x01) {$x_1$};
    \node[box, fill=gray!20, below=0cm of x01] (dots0) {$\vdots$};
    \node[box, fill=blue!20, below=0cm of dots0] (x0f) {$x_f$};

    \node[box, fill=red!20, below right = 0.2 and 0.2 of x00.north west] (x10) {$x_0$};
    \node[box, fill=orange!20, below=0cm of x10] (x11) {$x_1$};
    \node[box, fill=gray!20, below=0cm of x11] (dots1) {$\vdots$};
    \node[box, fill=blue!20, below=0cm of dots1] (x1f) {$x_f$};

    \node[box, fill=red!20, below right = 0.2 and 0.2 of x10.north west] (x0) {$x_0$};
    \node[box, fill=orange!20, below=0cm of x0] (x1) {$x_1$};
    \node[box, fill=gray!20, below=0cm of x1] (dots) {$\vdots$};
    \node[box, fill=blue!20, below=0cm of dots] (xf) {$x_f$};

    \node [below = \labelspacing of xf] {Training Data}; % Text label

    % -- Augmentation
    \node[box, above right=of dots.north east, fill=yellow!20] (aug) {$\psi(\cdot)$};
    \node [below = \labelspacing of aug] {Augmentation}; % Label

    \coordinate (aug_mid) at ($(aug.west)!0.5!(x1.east)$);
    \draw[-] (x1.east) -| (aug_mid.north);
    \draw[->] (aug_mid.south) |- (aug.west);

    % -- agumented samples
    \node[box, fill=gray!20, right =of aug.east, anchor = north west] (dotsa0) {$\vdots$};
    \node[box, fill=gray!20, below=0cm of dotsa0] (xa0f) {$\psi(x_f$)};
    \node[box, fill=gray!20, above=0cm of dotsa0] (xa01) {$\psi(x_1)$};
    \node[box, fill=gray!20, above =0cm of xa01] (xa00) {$\psi(x_0)$};

    \node[box, fill=gray!20, below right = 0.2 and 0.2 of xa00.north west] (xa10) {$x_0$};

    \node[box, fill=gray!20, below=0cm of xa10] (xa11) {$x_1$};
    \node[box, fill=gray!20, below=0cm of xa11] (dotsa1) {$\vdots$};
    \node[box, fill=gray!20, below=0cm of dotsa1] (xa1f) {$x_f$};

    \node[box, fill=gray!20, below right = 0.2 and 0.2 of xa10.north west] (xa0) {$\psi(x_0)$};
    \node[box, fill=gray!20, below=0cm of xa0] (xa1) {$\psi(x_1)$};
    \node[box, fill=gray!20, below=0cm of xa1] (dotsa) {$\vdots$};
    \node[box, fill=gray!20, below=0cm of dotsa] (xaf) {$\psi(x_f)$};

    \node [below = \labelspacing of xaf]{Augmented Data}; % Text label
    \draw[->] (aug.east) to (dotsa0.north west);
    
    % -- Neural Networks
    \node[box, right=of dotsa.north east, fill=yellow!20] (nn_aug) {$\phi_\theta(\cdot)$};
    \node [below = \labelspacing of nn_aug]{Neural Network}; % Label

    \node[coordinate, below right=of dots.north east] (theta_v_align) {};
    \node[box, anchor=north, fill=yellow!20] at (theta_v_align -| nn_aug.south) (theta) {$\phi_\theta(\cdot)$};

    \coordinate (theta_arrow_mark) at (aug_mid.south |- theta.west);
    \draw[-] (dots.east) -| (theta_arrow_mark.south);
    \draw[->] (theta_arrow_mark.north) |- (theta.west);

    %\node[box, below = of nn_aug.south, fill=yellow!20] (theta) {$\theta(\cdot)$};
    \node [below = \labelspacing of theta]{Neural Network}; % Label
    \draw[->] (dotsa.north east) to (nn_aug.west);

    % -- CLANP loss
    \coordinate (clanp_v_align) at ($(nn_aug.south east)!0.33!(nn_aug.north east)$);
    \node[box, right = of clanp_v_align.east, fill=yellow!20] (clanp_loss) {$L_{CLAN}$};
    \node [below = \labelspacing of clanp_loss] {SSL Loss}; % Label

    \coordinate (clanp_in_1) at ($(clanp_loss.south west)!0.67!(clanp_loss.north west)$);
    \draw[->] (nn_aug.east) to (clanp_in_1.west);

    \coordinate (clanp_in_2_start) at ($(theta.south east)!0.67!(theta.north east)$);
    \coordinate (clanp_in_2_end) at ($(clanp_loss.south west)!0.33!(clanp_loss.north west)$);
    \coordinate (clanp_in_2_mid) at ($(clanp_in_2_start)!0.5!(clanp_in_2_end)$);

    \draw[->] (clanp_in_2_mid.south) |- (clanp_in_2_end);
    \draw[-] (clanp_in_2_start) -| (clanp_in_2_mid.north);
    
    \coordinate (clanp_box) at ($(clanp_loss.south east) + (0,-0.75)$);
    \draw[dashed, thick] ($(clanp_loss.north west) + (-0.3, 0.2)$) rectangle ($(clanp_box) + (0.2, -0.2)$);

    \node[right = 0.5mm of clanp_in_2_mid, yshift = -5mm] {Training};

    % -- Finetune loss
    %\node[box, right=of nn_aug.east, fill=yellow!20] (ce_loss) {$L_{CE}$};
    %\node [below = \labelspacing of ce_loss]{Finetune Loss}; % Label
    
    % -- Centroid
    % calc
     
    \coordinate (theta_arrow_3) at ($(theta.south east)!0.33!(theta.north east)$);
    \node[box, right=of theta_arrow_3.east, fill=yellow!20] (centroid_calc) {$(\mu_0)_i = \frac{1}{N_{\text{train}}}\sum_{j=0}^{N_{\text{train}} -1} \phi_\theta(x_j)_i$};
    \node [below = \labelspacing of centroid_calc] (centroid_calc_label) {Centroid Calculation}; % Label
    \draw[->] (theta_arrow_3.east) -- (centroid_calc.west) node[midway, below=0.5mm] {Inference};

    % diagram
    \node[right= of centroid_calc] (centroid_diagram) {
        \resizebox{3cm}{!}{\input{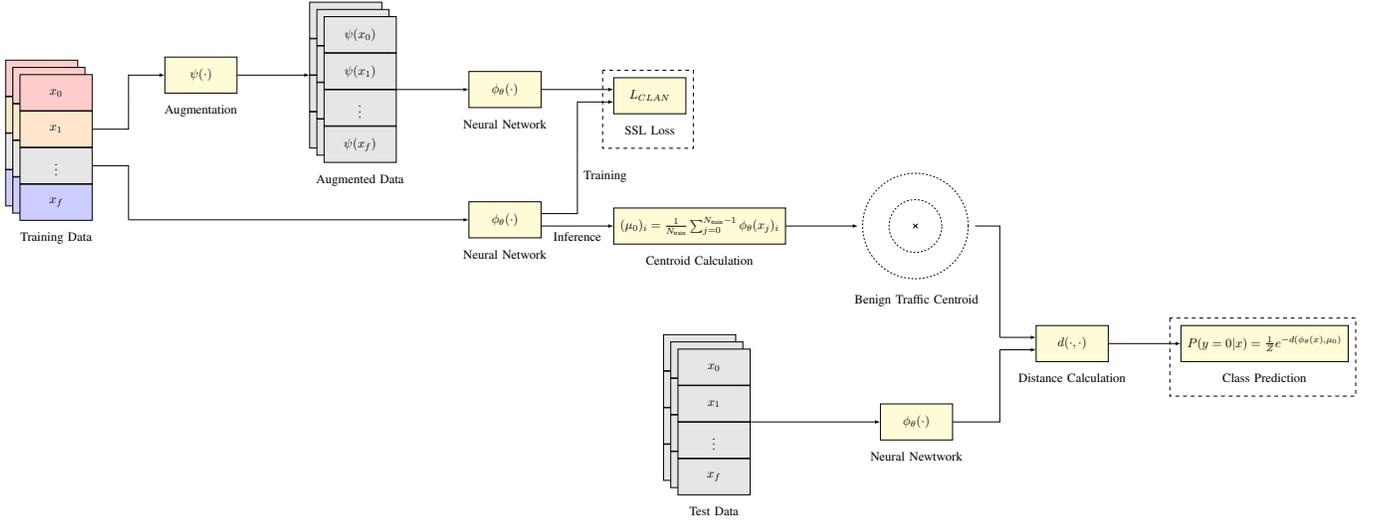}}
    };
    \node[below = \labelspacing of centroid_diagram]{Benign Traffic Centroid};
    \draw[->] (centroid_calc.east) to (centroid_diagram.west);

    % -- Test Data
    \node[box, fill=gray!20, below = 2.5cm of centroid_calc] (xv00) {$x_0$};
    \node[box, fill=gray!20, below=0cm of xv00] (xv01) {$x_1$};
    \node[box, fill=gray!20, below=0cm of xv01] (dots0) {$\vdots$};
    \node[box, fill=gray!20, below=0cm of dots0] (xv0f) {$x_f$};

    \node[box, fill=gray!20, below right = 0.2 and 0.2 of xv00.north west] (xv10) {$x_0$};
    \node[box, fill=gray!20, below=0cm of xv10] (xv11) {$x_1$};
    \node[box, fill=gray!20, below=0cm of xv11] (dotsv1) {$\vdots$};
    \node[box, fill=gray!20, below=0cm of dotsv1] (xv1f) {$x_f$};

    \node[box, fill=gray!20, below right = 0.2 and 0.2 of xv10.north west] (xv0) {$x_0$};
    \node[box, fill=gray!20, below=0cm of xv0] (xv1) {$x_1$};
    \node[box, fill=gray!20, below=0cm of xv1] (dotsv) {$\vdots$};
    \node[box, fill=gray!20, below=0cm of dotsv] (xvf) {$x_f$};

    \node[below = \labelspacing of xvf](val_input_text){Test Data};

    % -- Nerual Network
    \node[box,fill=yellow!20, anchor = center] (thetav) at  ([yshift=0cm]centroid_diagram.center|-dotsv.north east){$\phi_\theta(\cdot)$};
    \node[below = \labelspacing of thetav](nn_val_text){Neural Newtwork};
    \draw[->] (dotsv.north east) to (thetav.west);

   % -- Distance Calculation 
   \node[box, fill=yellow!20] (distance_calc) at ($(thetav.north east)!0.5!(centroid_diagram.south east) + (3,0)$) {$d(\cdot,\cdot)$};
   \node[below = \labelspacing of distance_calc]{Distance Calculation};

   \coordinate (d_arrow_int) at ($($(distance_calc.south west)!0.67!(distance_calc.north west)$) + (-1cm,0)$);
   \draw[-] (centroid_diagram.east) -| (d_arrow_int.south);
   \draw[->] (d_arrow_int.south) -- ($(distance_calc.south west)!0.67!(distance_calc.north west)$);

   \coordinate (d_arrow_int2) at ($(distance_calc.south west)!0.33!(distance_calc.north west) + (-1cm,0)$);
   \draw[-] (thetav.east) -| (d_arrow_int2.north);
   \draw[->] (d_arrow_int2.north) -- ($(distance_calc.south west)!0.33!(distance_calc.north west)$);

   % -- Inference
   \node[box, fill=yellow!20, right= of distance_calc] (inference) {$P(y=0|x) = \frac{1}{Z}e^{-d(\phi_\theta(x),\mu_0)}$};
   \node[below = \labelspacing of inference] {Class Prediction};
   \coordinate (inference_box) at ($(inference.south east) + (0,-0.75)$);
   \draw[->] (distance_calc.east) to (inference.west);
   \draw[dashed, thick] ($(inference.north west) + (-0.3, 0.2)$) rectangle ($(inference_box) + (0.2, -0.2)$);

\end{tikzpicture}
    }
    \caption{Overview of the CLAN framework. A neural network is trained on both genuine benign and augmented network traffic to learn the distribution of benign traffic and map it to a single distribution in latent space. Evaluating the probability of a test sample belonging to this distribution can then be used to infer its label during inference.}
    \label{fig:system_overview}
\end{figure*}

This section introduces the proposed Contrastive Learning using Augmented Negatives~(CLAN) framework. A wholistic overview of this approach is illustrated in Figure~\ref{fig:system_overview}. The CLAN loss function, described in Section~\ref{subsec:appraoch_loss}, provides the optimisation objective for training a neural network using a combination of benign traffic samples and their augmented variations. By minimising this loss, the model learns a latent representation where benign traffic forms a single latent distribution, from which other traffic types are distinctly separated. As detailed in Section~\ref{subsec:proposed_inference}, the centroid of this latent distribution can be computed and cached after training. During inference, a binary class label of an unknown test sample can then be inferred by evaluating the probability that it was sampled from the latent distribution based on the distance between the test sample's latent representation and this centroid.

\subsection{Contrastive Learning using Augmented Negatives}
\label{subsec:appraoch_loss}
The objective of self-supervised learning for network intrusion detection systems is to leverage a dataset of benign network traffic in order to learn a robust representation of the data, such that binary classification can be performed during inference. Concretely, given a set of possible class labels \( \mathcal{Y} := \{0,1,\ldots,N_C-1\}, \quad N_C \in \mathbb{Z}^+ \) and a training dataset containing only benign traffic, defined as \( \mathcal{D}_{\text{train}} = \{(x_i, y_i) \mid x_i \in R^f, y_i = 0, i = 1, \dots, N_{\text{train}}\}, N_{\text{train}}\in\mathbb{Z}^+ \) where \( y=0 \) corresponds to the benign class label and \( f\in\mathbb{Z}^+ \) represents the number of tabular features extracted for each sample, the conditional distribution of benign traffic must be learned such that a binary class label \( \hat{y} \in \{0,1\} \), corresponding to whether a test sample is predicted to be benign (\( y=0 \)) or malicious (\( y=1 \)), can be determined for samples of a test dataset defined as $\mathcal{D}_{\text{test}} = \{(x_i, y_i) \mid x_i \in R^f, y_i \in \mathcal{Y}, i = 1, \dots, N_{\text{test}}\}, N_{\text{test}}\in\mathbb{Z}^+$.

To achieve this the model must use the benign training examples to learn the conditional label distribution \( P(y=0|x) \). This can be expressed in terms of a joint probability \( P(y=0,x) \), and a partition function \( Z(x)\in\mathbb{R}^+ \) as shown in Equation~\ref{eq:conditional_dist}.

\begin{equation}\label{eq:conditional_dist}
  P(y=0|x) = \frac{P(x,y=0)}{Z(x)} = \frac{P(x, y=0)}{\sum_{c=0}^{N_C-1}P(x,y=c)}
\end{equation}

In this work a parameterised neural network \( \phi_\theta \) is used to map the input data to a latent representation \( z_i := \phi_\theta(x_i) \in \mathbb{R}^q \). It is assumed that each the class in the latent space follows a homoskedastic Gaussian distribution—i.e., each class \( i \) has a distinct mean \( \mu_i \) but shares the same isotropic covariance \( \sigma^2I \). Formally, the latent class distributions are modelled as \( P(z \mid y = i) = \mathcal{N}\bigl(z;\, \mu_i,\, \sigma^2 I\bigr) \). Under these assumptions the conditional probability $P(y=0|x)$ can be expressed as Equation~\ref{eq:parameterised_dist}, where \( d(\cdot,\cdot) \in \mathbb{R}^+ \) computes the squared Euclidean distance between two latent representations. 

\begin{equation}\label{eq:parameterised_dist}
  P(y=0|x) = \frac
  {e^{\frac{-d(z, \mu_0)}{2\sigma^2I}}}
  {e^{\frac{-d(z, \mu_\psi)}{2\sigma^2I}}}
\end{equation}

To learn a mapping from feature space to latent space, the negative log-likelihood of this distribution is minimised. After simplification, the negative log-likelihood across the training dataset is given by Equation~\ref{eq:simple_neg_log_likelihood}, where $z_a := \phi_\theta(x_a), \forall x_a \in \mathcal{D}_{\text{train}}$.

\begin{equation}\label{eq:simple_neg_log_likelihood}
  - \log L(\theta) \propto
  \sum_{a=0}^{N_{\text{train}-1}}[ 
    d(z_a, \mu_0)
    - \sum_{c=0}^{N_c-1} d(z_a, \mu_n)
  ]
  \end{equation}

However, at training time, the number of malicious classes—and consequently their latent mean vectors—is unknown. Additionally, computing the mean of the benign distribution dynamically during training is computationally expensive. To address this, the fact that the distance between a sample and the centroid of a Gaussian distribution is proportional to the expected distance between the sample and samples drawn from the Gaussian is exploited. This is stated formally in Equation~\ref{eq:gaussian_expectation} for a latent vector \( z \) and a second latent vector \( z' \) drawn from class \emph{i}. This allows for the distance between a latent representation and a distribution centroid to be estimated via Monte Carlo sampling.

\begin{equation}\label{eq:gaussian_expectation}
  d(z,\mu_i) \propto \mathbb{E}_{z' \sim \mathcal{N}(\mu_i, \sigma^2 I)}\bigl[d(z,z')\bigr]
\end{equation}

Since malicious samples are unavailable during training, a surrogate distribution \( \psi(x) \in \mathbb{R}^f \) is used instead. Specifically, this surrogate is constructed by resampling features of \( x \) uniformly within the range \( [-b,b] \) with probability \( p_{\text{resample}} \in (0,1] \), where \( b \in \mathbb{R}^+ \) and \( p_{\text{resample}} \) are hyperparameters. It assumed that an even number of samples of each class, both benign and malicious, are drawn from this distribution. Substituting Monte Carlo distance estimation into Equation~\ref{eq:simple_neg_log_likelihood} gives the negative log-likelihood expression shown in Equation~\ref{eq:monte_carlo_estimation}. Here \( \omega_a := \phi_\theta(x_a) \) and \( \tilde{\omega}_a := \phi_\theta(\psi(x_a)) \) are the latent representations of original and augmented samples, respectively, for a set \( \mathcal{S} = \{x_i\}_{i=0}^k \) of \( k \) samples drawn from \( \mathcal{D}_{\text{train}} \).

\begin{equation}\label{eq:monte_carlo_estimation}
  - \log L(\theta) \propto
  \sum_{a=0}^{N_{\text{train}-1}} [\sum_{p=0}^{k-1} d(z_a, \omega_p) - \sum_{n=0}^{k-1} d(z_a, \tilde{\omega}_n)]
\end{equation}

Finally, calculating this over a batch, \( x \in \mathbb{R}^{B \times f} \), and introducing a hinge regularisation term with margin hyperparameter \( m \in (0,1] \) results in the proposed CLAN loss function given in Equation~\ref{eq:clan_loss}.

\begin{align}\label{eq:clan_loss}
  L_{CLAN}(x) &=  
   \frac{1}{B}[
    \sum_{a=0}^{B-1}[
      \sum_{p=0 \atop p \neq a}^{B-1} d(\phi_\theta(x_a), \phi_\theta(x_p)) \notag\\
      &+ \sum_{n=0}^{B-1} \max(0, m - d(\phi_\theta(x_a), \phi_\theta(\psi(x_n))))
   ]]
\end{align}
  
The CLAN loss function jointly optimises the latent space to model benign traffic as a Gaussian distribution while simultaneously performing maximum likelihood estimation to learn the mapping from feature space to latent space. This results in a representation where benign traffic is clustered around a centroid, while malicious traffic is pushed away from this cluster. It can be shown that using the cosine distance metric optimises a similar objective while replacing the Gaussian assumption with a von Mises-Fisher distribution assumption. This was found to improve performance and is thus used in the experiments in Section~\ref{sec:results}.

\subsection{Probabilistic Inference}
\label{subsec:proposed_inference}

CLAN models benign traffic as a distribution in latent space. Analysing the loss function reveals that the unnormalised joint probability of sampling both the input sample and the benign distribution exhibits an exponential decay with respect to the distance metric optimised and the distribution's centroid, as expressed in Equation~\ref{eq:unnorm_benign_joint}. Here \( d(\cdot,\cdot) \in \mathbb{R}^+ \) represents the distance metric optimised by the loss function.

\begin{equation}\label{eq:unnorm_benign_joint}
  \tilde{P}(x,y=0) = e^{-d(\phi_\theta(x),\mu_0)}
\end{equation}

During inference, the model parameters remain fixed, allowing the centroid of the benign traffic distribution to be precomputed as the geometric mean of latent representations in the training dataset, as shown in Equation~\ref{eq:centroid_calc}.

\begin{equation}\label{eq:centroid_calc}
  \mu_0 = \frac{1}{N_{\text{train}}}\sum_{x \in \mathcal{D_{\text{train}}}}\phi_\theta(x)
\end{equation}

The conditional probability of a test sample being benign is then computed by normalising the distance between its latent representation and the benign centroid using a partition function, as defined in Equation~\ref{eq:scaled_exp_decay}.

\begin{equation} \label{eq:scaled_exp_decay}
  P(y=0|x) = \frac{1}{Z}e^{-d(\phi(x),\mu_0)}
\end{equation}

Here, the partition function, \( Z \in \mathbb{R}^+ \), is treated as a constant hyperparameter that controls the trade-off between false positive rate and recall. The final classification decision assigns a predicted label $\hat{y} \in \{0,1\}$ based on a probability threshold: a test sample is classified as benign (\(\hat{y} = 0\)) if \(P(y=0|x) > 0.5\), and as malicious (\(\hat{y} = 1\)) otherwise, as defined in Equation~\ref{eq:ssl_pred_label}.

\begin{equation}
  \hat{y} = 
  \begin{cases} 
      0, & \text{if } P(y=0|x) > 0.5; \\
      1, & \text{otherwise}.
  \end{cases}
  \label{eq:ssl_pred_label}
\end{equation}

\section{Comparison to Existing SSL Approaches}
\label{sec:contrastive_comparison}

The proposed loss function follows the same general form as existing contrastive loss functions and can be described under the unified contrastive loss framework by setting specific parameters and incorporating the hinge regularisation term~\cite{DBLP:journals/corr/abs-2102-06810}, which was included in the original contrastive loss function formulation~\cite{contrastive_loss2,constrastive_loss1}. Furthermore, by removing the hinge function, the CLAN loss function becomes equivalent to the NTXent loss~\cite{NIPS2016_6b180037} with a temperature value of one. Existing SSL approaches for NIDS implement similar methodologies. For example, SSCL-IDS~\cite{10619725}, CLDNN~\cite{9935282}, and Conflow~\cite{conflow} all optimise a loss of the form shown in Equation~\ref{eq:existing_ssl_info_nce}, where \( \psi(x) \) represents an arbitrary augmentation function, and \( \tau \in \mathbb{R}^+ \) is a temperature hyperparameter.

\begin{equation}
  L(x) = - \frac{1}{B} \sum_{a=0}^{B-1}\log (
    \frac{e^{\frac{-d(\phi(x_a),\phi(\psi(x_a)))}{\tau}}}
    {\sum_{j=0}^{B-1} e^{\frac{-d(\phi(x_a), \phi(x_j))}{\tau}}}
    )
    \label{eq:existing_ssl_info_nce}
\end{equation}

\begin{figure}[t]
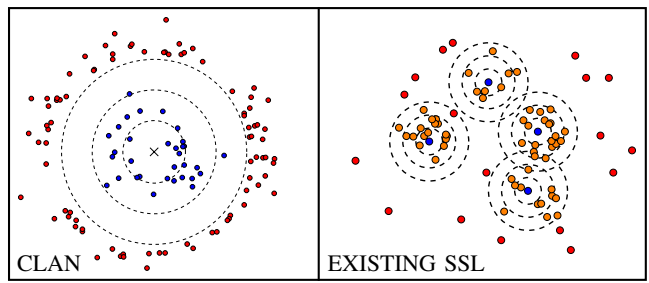

    \centering
    \resizebox{\linewidth}{!}{
        \begin{tikzpicture}
    \usetikzlibrary{fit,positioning}

    % -- Top figure (Anomaly Detection)
    \node[anchor=east, inner sep=0pt] (clanp) {
        \resizebox{0.4\linewidth}{!}{\input{diagrams/embedding_comp/clanp_embeddings_tikz}}
    };  
        
    % -- Bottom figure (Open Set Recognition), positioned to the right
    \node[anchor=west] (ssl) at ([xshift=1cm]clanp.east) {
        \resizebox{0.4\linewidth}{!}{\input{diagrams/embedding_comp/ssl_embedding_tikz}}
    };

    % Define a bounding box that encompasses both figures
    \node[fit=(clanp) (ssl), draw, thick] (boundingBox) {};

    % Define the midpoint coordinate horizontally centered in boundingBox
    \path (clanp.east) ++(0.5cm, 0) coordinate (mid);

    % Draw vertical separating line at the middle
    \draw[thick] (mid |- boundingBox.north) -- (mid |- boundingBox.south);

    % -- Add title to the top-left box (CLAN)
    \node[anchor=south west, font=\small] at ($(boundingBox.south west)$) {CLAN};

    % -- Add title to the bottom-right box (EXISTING SSL)
    \node[anchor=south west, font=\small] at ($(mid |- boundingBox.south)$) {EXISTING SSL};

\end{tikzpicture}
    }
    \caption{Comparison of the latent representations learned by the CLAN loss function to those learned by existing self-supervised loss functions. \textbf{Left:} CLAN learns a single distribution corresponding to benign traffic (blue) whilst malicious traffic (red) appears outwith this distribution. \textbf{Right:} Existing self-supervised loss functions learn a distribution for each benign sample and its augmented views (orange).}
    \label{fig:embedding_comp}
    
\end{figure}

The key distinction between CLAN and existing approaches lies in how augmented samples are treated. While CLAN considers augmented samples as negative pairs, existing approaches treat them as positive pairs. This difference arises from the underlying assumptions in the derivation of their respective loss functions. CLAN performs maximum likelihood estimation under the assumption that benign traffic forms a single distribution in the latent space. In contrast, existing approaches can be derived following CLAN's derivation, except under the assumption that each sample and its augmented versions each form a distinct distribution in latent space. This distinction is illustrated in Figure~\ref{fig:embedding_comp}, assuming that the loss functions optimise the squared Euclidean distance under the Gaussian assumption.

CLAN learning a single latent distribution of benign traffic provides several advantages over other SSL methods, which model the latent space as a mixture of distributions. Firstly, CLAN directly learns the overall distribution of benign network traffic, whereas existing SSL methods attempt to improve robustness by mapping noise-induced variations from augmentation to a structured representation. This distinction results in significant performance improvements in both binary classification and fine-tuned performance, as demonstrated in Section~\ref{sec:results}. Notably, this approach is only viable under the assumption that the majority of samples belong to a single class distribution~(benign traffic), which has not been exploited by prior SSL methods primarily developed for image-based tasks.

CLAN also offers advantages in computational efficiency during inference. Since classification is performed by evaluating the probability of a test sample belonging to the latent benign distribution, only a single distance measurement between the test representation and the centroid is required. Assuming the centroid of the training dataset representations is computed and cached post-training, inference incurs a fixed computational cost, resulting in a complexity of \( \mathcal{O}(1) \). In contrast, existing SSL approaches learn a separate distribution for each training sample. During inference, these methods evaluate the probability of a test sample belonging to the nearest learned distribution by thresholding the nearest-neighbour distance. However, identifying the closest distribution requires a nearest-neighbour search, leading to a computational complexity of \( \mathcal{O}(N_{\text{train}}) \). Consequently, the CLAN framework is significantly more scalable to the demands of modern network NIDS, which may need to monitor millions of flows per day.

\section{Experimental Results}
\label{sec:results}

This section experimentally compares the CLAN framework to existing approaches in literature. Initially, the experimental procedure is described in Section~\ref{subsec:results_procedure}. The effectiveness of CLAN is then evaluated by comparing its performance in binary classification to existing SSL approaches in Section~\ref{subsec:results_baseilnes}, and to anomaly detectors in Section~\ref{subsec:results_anomaly_detection}. Finally, the performance of CLAN is evaluated when fine-tuned on a limited datset to perform multi-class classification in Section~\ref{subsec:results_multiclass_classification}.
 
\subsection{Experimental Procedure}
\label{subsec:results_procedure}
In this work, the CLAN loss function was used to train a modified MLP architecture. The architecture begins with a linear transformation that projects the input features from \( \mathbb{R}^f \) to \( \mathbb{R}^{d_{\text{model}}} \), where \( d_{\text{model}} \in \mathbb{Z}^+ \). This is followed by a sequence of fully connected layers, each followed by ReLU activation functions. A final linear transformation projects the data down to \( \mathbb{R}^{d_{\text{head}}} \) where \( d_{\text{head}} \in \mathbb{Z}^+\) such that \( d_{\text{head}} < d_{\text{model}}\). The network’s width and depth were treated as hyperparameters and optimised accordingly. Additionally, the uniform resampling rate and range used in CLAN were treated as hyperparameters. Baseline models were trained using the architectures specified in their respective original implementations.

To evaluate the effectiveness of CLAN compared to baseline models, models were trained and tested on the Lycos2017 dataset~\cite{lycos2017}, an improved version of the CICIDS2017 dataset~\cite{cicids2017} that addresses various feature extraction and labelling errors. The dataset consists of 1,789,954 network flows across 14 classes, including benign traffic. It is highly imbalanced, with benign traffic accounting for over 1,000,000 samples, while certain malicious classes contain as few as 11 samples. 

The dataset was partitioned train and test splits using a stratified sampling approach. Specifically, for each class, 50\% of the available samples were randomly selected for the training set, and the remaining 50\% were assigned to the test set: ensuring balanced representation across both splits. Exceptions were made for the Heartbleed and SQL Injection classes, which were included exclusively in the test set due to their limited sample size.

In the binary classification comparisons given in Section~\ref{subsec:results_baseilnes} and Section~\ref{subsec:results_anomaly_detection} models were optimised using 200 iterations of random search, with each iteration employing 5-fold stratified cross-validation, with malicious traffic being discarded from each training partition. The best performing configuration was subsequently retrained on benign traffic across the entire training dataset and evaluated on the test set. Models were trained for 200 epochs using the AdamW optimizer with a warm-up cosine learning rate schedule. The base learning rate, batch size, weight decay, and model-specific parameters were treated as hyperparameters.

% finetune comparison
In the fine-tuned performance comparisons given in Section~\ref{subsec:results_multiclass_classification} the weights learned during pretraining in Section~\ref{subsec:results_baseilnes} were fine-tuned on a limited subset of the training data which was generated through stratified sampling for 100 epochs using the AdamW optimiser with a learning rate of \(10^{-6}\) and batch size of 64. Due to the limited size of the fine-tuning dataset, reported results were averaged over 10 runs, with a different seed being used to sample the subset of training data each time.

\subsection{Comparison to SSL Approaches}
\label{subsec:results_baseilnes}

In this section, CLAN is pretrained on benign network traffic and deployed as a binary classifier without fine-tuning. Several existing contrastive learning approaches were selected as baselines, including SSCL-IDS~\cite{10619725}, CLDNN~\cite{9935282}, and Conflow~\cite{conflow}, each of which employs distinct augmentation strategies. Additionally, BYOL, SimSiam, Barlow Twins, and VICReg were chosen as non-contrastive SSL baselines and trained using the model architectures and training protocols outlined in previous works~\cite{non_contrastive_ssl}. 

The class-wise AUROC scores, along with the mean AUROC for each model, are presented in Table~\ref{table:ssl_auroc}. The results demonstrate the effectiveness of CLAN, which achieves significant performance improvements over existing SSL models by learning a holistic representation of benign traffic. In contrast, conventional SSL approaches model a separate latent distribution for each individual sample, limiting their ability to generalise effectively in a network intrusion detection setting.

\begin{table*}
    \caption{AUROC comparison of CLAN and existing SSL baselines when performing binary classification without fine-tuning.}
    \centering
    \resizebox{\linewidth}{!}{
    \begin{tabular}{lccccccccccccc}
      \toprule
      Class & CLAN & CLDNN~\cite{9935282} & SSCL-IDS~\cite{10619725} & ConFlow~\cite{conflow} & Barlow Twins~\cite{non_contrastive_ssl} & SimSiam~\cite{non_contrastive_ssl} & BYOL~\cite{non_contrastive_ssl,byol2} & VICReg~\cite{non_contrastive_ssl} \\
      \midrule
      Botnet & 0.915536 & 0.951724 & 0.953530 & 0.927866 & 0.977046 & 0.996176 & 0.985893 & 0.946107 \\
      DDoS & 0.996584 & 0.989344 & 0.972851 & 0.925385 & 0.999179 & 0.999191 & 0.999782 & 0.999400 \\
      DoS (Golden Eye) & 0.931653 & 0.979910 & 0.841126 & 0.732904 & 0.896085 & 0.899778 & 0.905229 & 0.924664 \\
      DoS (Hulk) & 0.977893 & 0.979486 & 0.982326 & 0.910926 & 0.994941 & 0.997693 & 0.990773 & 0.997170 \\
      DoS (Slow HTTP Test) & 0.991794 & 0.613476 & 0.523339 & 0.522115 & 0.633869 & 0.542329 & 0.508903 & 0.522023 \\
      DoS (Slow Loris) & 0.992513 & 0.873563 & 0.966009 & 0.953790 & 0.990351 & 0.982546 & 0.995689 & 0.984311 \\
      FTP Patator & 0.944212 & 0.996560 & 0.996554 & 0.997778 & 0.993760 & 0.999653 & 0.998454 & 0.996132 \\
      Portscan & 0.989271 & 0.849846 & 0.994321 & 0.993277 & 0.979313 & 0.990024 & 0.997439 & 0.985080 \\
      SSH Patator (Brute Force) & 0.956259 & 0.958037 & 0.784271 & 0.903715 & 0.914556 & 0.864209 & 0.905103 & 0.840080 \\
      Web Attack (Brute Force) & 0.907790 & 0.784891 & 0.721126 & 0.817545 & 0.678314 & 0.786032 & 0.755342 & 0.654605 \\
      Web Attack (XSS) & 0.963409 & 0.849181 & 0.800105 & 0.880714 & 0.753717 & 0.765806 & 0.728249 & 0.690359 \\
      Heartbleed & 0.997604 & 0.942537 & 0.993278 & 0.988757 & 0.999930 & 0.999278 & 0.999655 & 0.999871 \\
      Web Attack (SQL Injection) & 0.897170 & 0.948995 & 0.966933 & 0.910350 & 0.916327 & 0.963398 & 0.954639 & 0.951767 \\
      \textbf{Mean} & \textbf{0.958591} & 0.901350 & 0.884290 & 0.881933 & 0.902107 & 0.906624 & 0.901935 & 0.883967 \\
      \bottomrule
    \end{tabular}
    }
    \label{table:ssl_auroc}
  \end{table*}

\subsection{Comparison to Anomaly Detectors}
\label{subsec:results_anomaly_detection}

Next, CLAN was compared against several baseline anomaly detection methods under the same evaluation setting as the SSL experiments. The baselines were selected to represent a range of approaches: gradient-free (Isolation Forrest~\cite{10040395} and SVM~\cite{Hanan_svm1}), deep reconstruction (Autoencoder~\cite{10.1007/978-3-319-13563-2_27}, DAE-LR~\cite{nkashama2024deeplearningnetworkanomaly}, DAGMM~\cite{purohit2020deepautoencodinggmmbasedunsupervised} and DUAD~\cite{li2021deep}), and deep one-class learning (autoSVM~\cite{8463474} and Deep SVDD~\cite{pmlr-v80-ruff18a}). The AUROC scores for each baseline are reported in Table~\ref{tab:anomaly_detection}. The results further demonstrate the effectiveness of CLAN, which significantly outperforms all anomaly detection baselines.

  \begin{table*}
    \caption{AUROC comparison of CLAN and existing anomaly detectors when performing binary classification. Columns ordered by mean performance.}
    \centering
    \resizebox{\linewidth}{!}{
    \begin{tabular}{lcccccccccc}
      \toprule
      Class & CLAN & DUAD~\cite{li2021deep} & DAE-LR~\cite{nkashama2024deeplearningnetworkanomaly} & Deep SVDD~\cite{pmlr-v80-ruff18a} & AE~\cite{10.1007/978-3-319-13563-2_27} & IF~\cite{10040395} & AutoSVM~\cite{8463474} & SVM~\cite{Hanan_svm1} & DAGMM~\cite{purohit2020deepautoencodinggmmbasedunsupervised} \\
      \midrule
      
      Botnet & 0.915536 & 0.819026 & 0.759901 & 0.751007 & 0.671062 & 0.629887 & 0.641687 & 0.637679 & 0.605228 \\
      DDoS & 0.996584 & 0.979402 & 0.997996 & 0.996513 & 0.908072 & 0.945021 & 0.933722 & 0.889780 & 0.876049 \\
      DoS (Golden Eye) & 0.931653 & 0.951141 & 0.952424 & 0.892923 & 0.855621 & 0.923275 & 0.887754 & 0.846840 & 0.736878 \\
      DoS (Hulk) & 0.977893 & 0.967905 & 0.994860 & 0.993388 & 0.906273 & 0.955478 & 0.926223 & 0.894898 & 0.589771 \\
      DoS (Slow HTTP Test) & 0.991794 & 0.954526 & 0.981080 & 0.975533 & 0.962552 & 0.966122 & 0.968576 & 0.963021 & 0.899879 \\
      DoS (Slow Loris) & 0.992513 & 0.929777 & 0.986823 & 0.970622 & 0.897970 & 0.904747 & 0.880789 & 0.896824 & 0.955010 \\
      FTP Patator & 0.944212 & 0.962427 & 0.959192 & 0.968324 & 0.765372 & 0.762553 & 0.753264 & 0.736828 & 0.748862 \\
      Portscan & 0.989271 & 0.982759 & 0.983505 & 0.948716 & 0.720184 & 0.803987 & 0.668389 & 0.741269 & 0.517609 \\
      SSH Patator (Brute Force) & 0.956259 & 0.889312 & 0.964803 & 0.962553 & 0.857118 & 0.845189 & 0.874968 & 0.799072 & 0.887558 \\
      Web Attack (Brute Force) & 0.907790 & 0.864899 & 0.594903 & 0.673249 & 0.784379 & 0.713374 & 0.736061 & 0.767771 & 0.825337 \\
      Web Attack (XSS) & 0.963409 & 0.942706 & 0.564499 & 0.658500 & 0.780935 & 0.702785 & 0.730008 & 0.761677 & 0.810560 \\
      Heartbleed & 0.997604 & 0.979357 & 0.999801 & 0.998676 & 0.989079 & 0.999801 & 0.994511 & 0.993468 & 0.978630 \\
      Web Attack (SQL Injection) & 0.897170 & 0.830627 & 0.848514 & 0.744588 & 0.731559 & 0.752289 & 0.745230 & 0.745087 & 0.653157 \\
      \textbf{Mean} & \textbf{0.958591} & 0.927221 & 0.891408 & 0.887276 & 0.833090 & 0.838808 & 0.826245 & 0.821093 & 0.775733 \\
      
      \bottomrule
    \end{tabular}
    }
    \label{tab:anomaly_detection}
\end{table*}

\subsection{Fine-tuning Comparison}
\label{subsec:results_multiclass_classification}

Finally, to assess the effectiveness of CLAN’s learned representations for multi-class classification, a linear layer was appended to each SSL model, which was then fine-tuned using a limited number of samples per class. The mean macro-averaged F1 scores for each model, averaged over 10 runs for each training set size, are presented in Table~\ref{tab:finetune_results}. 

\begin{table*}
    \caption{Macro averaged F1 scores of CLAN and SSL baselines when fine-tuned on a limited dataset to perform multi-class classification.}
    \centering
    \resizebox{\linewidth}{!}{
    \begin{tabular}{lcccccccc}
      \toprule
      Samples & CLAN & CLDNN \cite{9935282} & SSCL-IDS \cite{10619725} & Conflow \cite{conflow} & Barlow Twins \cite{non_contrastive_ssl} & SimSiam \cite{non_contrastive_ssl} & BYOL \cite{non_contrastive_ssl,byol2} & VICReg \cite{non_contrastive_ssl} \\
      \midrule
      8   & \textbf{0.496316} & 0.379488 & 0.469348 & 0.480401 & 0.361244 & 0.482876 & 0.460014 & 0.389542 \\
      16  & \textbf{0.538254} & 0.438815 & 0.509693 & 0.501118 & 0.455801 & 0.470977 & 0.521448 & 0.440088 \\
      32  & \textbf{0.544964} & 0.483641 & 0.532057 & 0.532863 & 0.490422 & 0.512560 & 0.540308 & 0.513705 \\
      64  & \textbf{0.589188} & 0.535440 & 0.574097 & 0.571705 & 0.534474 & 0.555894 & 0.582854 & 0.528189 \\
      128 & \textbf{0.628799} & 0.589808 & 0.612588 & 0.617402 & 0.581268 & 0.603954 & 0.626497 & 0.593119 \\
      256 & 0.655416 & 0.644582 & 0.650703 & 0.651041 & 0.634673 & 0.645438 & \textbf{0.657744} & 0.631990 \\
      512 & \textbf{0.710183} & 0.683213 & 0.693791 & 0.705925 & 0.669987 & 0.684013 & 0.703609 & 0.665208 \\
      1024 & \textbf{0.738838} & 0.729141 & 0.731739 & 0.735170 & 0.719939 & 0.726593 & 0.735062 & 0.728573 \\
      \bottomrule
    \end{tabular}
    }
    \label{tab:finetune_results}
  \end{table*} 

The results demonstrate that CLAN outperforms all baseline models across all training set sizes, except for the case where the training set consists of 256 samples per class, where it is marginally outperformed by BYOL. However, this appears to be an isolated occurrence, as CLAN consistently outperforms BYOL on all other training set sizes. These findings underscore the effectiveness of CLAN in learning robust priors that facilitate fine-tuning for downstream classification tasks.

\section{Discussion and Conclusions}
\label{sec:conclusions}

This work introduced Contrastive Learning using Augmented Negative Pairs~(CLAN), a novel self-supervised learning framework for network intrusion detection systems. Unlike conventional contrastive learning approaches that treat augmented views as positive pairs, CLAN treats augmented samples as negative pairs, belonging to a potentially malicious distribution. While existing SSL approaches learn a distinct latent distribution for each training sample, CLAN instead learns a single cohesive distribution of benign network traffic. This paradigm shift results in improved performance and computational efficiency when applied to both anomaly detection and supervised classification tasks.

Through experimental evaluation on the Lycos2017 dataset, CLAN was compared to existing approaches in anomaly detection and self-supervised learning in a binary classification task, where it was found to outperform the leading approaches by an AUROC improvement of 0.031370 and 0.056484 respectively. Additionally, when fine-tuned with a limited quantity of labelled samples, CLAN demonstrated improved multiclass classification performance over existing self-supervised learning models, highlighting its effectiveness in real-world scenarios where labelled data is scarce.

Beyond its performance benefits, CLAN also offers advantages in computational efficiency. By modelling benign traffic as a single distribution, inference requires only one distance measurement giving a resultant complexity of \( \mathcal{O}(1) \), making it highly scalable for large-scale deployments. In contrast, existing SSL approaches require a nearest-neighbour search over training samples, leading to a complexity of \( \mathcal{O}(N_{\text{train}}) \), which becomes impractical in high-throughput network environments.

One limitation of the current approach is that it assumes that all training data is benign, which may not hold in real-world environments where the training data may be polluted with malicious samples. In future work, CLAN's robustness will evaluated under such conditions. Overall, CLAN represents a significant step forward in self-supervised learning for network intrusion detection systems, providing more efficient and effective classification. It is hoped that future works will build upon the CLAN framework by also treating augmented samples as negative pairs when training self-supervised models.
 
\bibliographystyle{IEEEtran}
\bibliography{IEEEabrv, references}

\end{document}